\documentclass[a4paper,fleqn]{cas-dc}

\usepackage[authoryear,longnamesfirst]{natbib}
\usepackage{caption}
\usepackage{subcaption}
\usepackage{graphicx}
\usepackage{siunitx}
\usepackage{enumitem}
\usepackage{amsmath}
\usepackage{multirow}
\usepackage{xurl}
\usepackage{float}
\usepackage{color,soul}
\usepackage{etoolbox}
\newcommand{\hlc}[1]{\sethlcolor{white}\hl{#1}}%
\newcommand{\hlcc}[1]{\sethlcolor{white}\hl{#1}}%
\DeclareMathOperator{\argmin}{argmin}

\def\tsc#1{\csdef{#1}{\textsc{\lowercase{#1}}\xspace}}
\tsc{WGM}
\tsc{QE}
\tsc{EP}
\tsc{PMS}
\tsc{BEC}
\tsc{DE}

\makeatletter
\patchcmd{\ps@pprintTitle}
  {Preprint submitted}
  {To be submitted}
  {}{}
\makeatother
\begin{document}
\let\WriteBookmarks\relax
\def\floatpagepagefraction{1}
\def\textpagefraction{.001}

\shorttitle{BeamsNet: A data-driven Approach Enhancing \hlc{Doppler Velocity Log} Measurements for Autonomous Underwater Vehicle Navigation}

\shortauthors{N.Cohen and I.Klein}

\title [mode = title]{BeamsNet: A data-driven Approach Enhancing Doppler Velocity Log Measurements for Autonomous Underwater Vehicle Navigation}



%
\author{Nadav Cohen}[orcid=0000-0002-8249-0239]

\cormark[1]


\ead{ncohe140@campus.haifa.ac.il}



\address{The Hatter Department of Marine Technologies, University of Haifa, Haifa, Israel}

\author{Itzik Klein}[orcid=0000-0001-7846-0654]

\cortext[cor1]{Corresponding author}
\begin{abstract}
Autonomous underwater vehicles (AUV) perform various applications such as seafloor mapping and underwater structure health monitoring. 
Commonly, an inertial navigation system aided by a Doppler velocity log (DVL) is used to provide the vehicle's navigation solution. In such fusion, the DVL provides the velocity vector of the AUV, which determines the navigation solution's accuracy and helps estimate the navigation states. 
This paper proposes BeamsNet, an end-to-end deep learning framework to regress the estimated DVL velocity vector that improves the accuracy of the velocity vector estimate, \hlc{and could replace the model-based approach}. Two versions of BeamsNet, differing in their input to the network, are suggested. The first uses the current DVL beam measurements and inertial sensors data, while the other utilizes only DVL data, taking the current and past DVL measurements for the regression process. Both simulation and sea experiments were made to validate the proposed learning approach relative to the model-based approach. Sea experiments were made with the Snapir AUV in the Mediterranean Sea, collecting approximately four hours of DVL and inertial sensor data.  
Our results show that the proposed approach achieved an improvement of more than 60\% in estimating the DVL velocity vector.

\end{abstract}




\maketitle

\section{Introduction}

An autonomous underwater vehicle (AUV) is a robotic platform capable of self-propulsion in an underwater environment. The AUV is able to maneuver underwater in three dimensions and is usually controlled by an onboard computer. For the purpose of operating autonomously, it contains several sensors collecting data enabling navigation \cite{jain2015review}.  AUVs are used in various fields such as oceanographic survey and mapping, undersea oil and gas exploration, ship hull inspection, and military applications \cite{nicholson2008present,manalang2018resident}.   \\
Autonomous navigation is a crucial aspect of the AUV operation, not only because it needs to operate in places beyond human reach and return safely, but also because when an area is explored, the specific location is essential information \cite{leonard2016autonomous}. The global navigation satellite systems (GNSS) receiver provides high positioning accuracy in many navigation solutions. However, in an underwater environment, the GNSS fails to receive the satellite signals and cannot be used \cite{liu2018innovative}. For that reason, underwater navigation and localization techniques were researched in the fields of inertial/dead reckoning positioning based on acoustic beacons and modems and geophysical navigation, which aims to use sensor measurements of geophysical parameters or environmental features to estimate the position of the AUV \cite{paull2013auv}. \\
A highly promising solution for the navigation problem in AUVs integrates an inertial navigation system (INS) and Doppler velocity log (DVL) sensors. An INS has a computer to calculate the navigation solution based on its inertial sensor readings located in the inertial measurement unit (IMU). It consists of a three-axis accelerometer and a three-axis gyroscope, which provides the AUV's specific force and angular velocity vectors \cite{titterton2004strapdown,ahmad2013reviews}. 
By using these measurements and solving the INS equations of motion, one can obtain the platform's position, velocity, and orientation \cite{groves2015principles,shin2002accuracy}. 
However, in a real-life application, the IMU is subject to errors such as misalignment between the sensor, bias, noise, and more, which makes the navigation solution error propagate over time. Therefore, tracking an AUV based on the INS readings alone is unrealistic \cite{thong2002dependence,akeila2013reducing}.\\
The DVL sensor is built from four transducers that emit four acoustic beams to the seafloor. The main configuration of the transducers is "$\times$", which can be seen in Figure\ref{fig1}, and indicates that the four beams are horizontally orthogonal. Once the beams are reflected back to the sensor from the seafloor, the AUV's velocity can be estimated. The DVL is considered an accurate sensor for velocity measurements and can achieve a velocity measurement accuracy of 0.2\% of the current velocity \cite{liu2018ins,wang2019novel}. For this reason, the INS is commonly fused with the DVL, and different integration methods have been researched. A common approach is to use nonlinear estimation such as an extended Kalman filter (EKF) or unscented Kalman filter (UKF) with the DVL as an aiding sensor \cite{zhang2018novel,zhang2019application,liu2021modified}. Other aspects of such fusion address the alignment and calibration between the INS and DVL. For example, \cite{li2015alignment} an alignment calibration of IMU and DVL was presented for better precision. In all of the methods above, the DVL, due to its accuracy, is used to determine the fusion accuracy.\\
In parallel to the developments in underwater navigation, data-driven approaches show great results in different fields to improve navigation accuracy and robustness. In \cite{shurin2022autonomous} deep hybrid learning approach was implemented to improve quad-rotor dead reckoning. In the field of indoor navigation with pedestrian dead reckoning, learning frameworks showed superior results over model-based approaches  \cite{gu2018accurate,chen2020deep,asraf2021pdrnet}. Furthermore, indoor robot navigation was improved using data-driven methods such as deep reinforcement learning \cite{zhu2017target,hu2021sim}.\\
In addition, data-driven approaches using DVL measurements, \hlcc{operating in normal conditions,} have been researched in recent years. A deep learning network called "NavNet" that combines the data from attitude and heading reference system (AHRS) and DVL was proposed in \cite{zhang2020navnet} and showed good performance in terms of both navigation accuracy and fault tolerance. Furthermore, \cite{mu2019end} suggested end-to-end navigation with AHRS and DVL with \hlcc{hybrid recurrent neural networks and} \cite{topini2020lstm}\hlc{ proposed a long short term memory based dead-reckoning approach that estimates the AUV surge and sway velocities by using temporal sequences of generalized forces and past estimated AUV velocities. } \hlcc{Later,} \cite{lv2021position} \hlc{ put forward a hybrid gated recurrent neural network for position correction model that, unlike standard navigation algorithms, does not require a motion model in order to avoid modeling errors in the navigation process} and \cite{liu2022sins} \hlc{ demonstrated an INS/DVL integration method based on a radial basis function neural network for current compensation}. \hlcc{In other works, scenarios where the DVL is not operating in normal conditions, such DVL malfunction, partial beam measurements or outliers, were addressed with machine and deep learning solutions }\cite{yona2021compensating,davari2021real,li2021underwater,saksvik2021deep,lv2020underwater}.

A recently published work used a deep learning method to solve the bearing-only localization problem instead of using the model-based iterative least squares estimator \cite{shalev2021botnet}. They showed, by simulation, that a data-driven deep learning approach performs better than the iterative least squares. Although \cite{shalev2021botnet} working on a nonlinear problem with different sensors and a localization problem instead of a navigation problem, it gives a good indication that deep learning can obtain better results compared to a standard parameter estimator such as LS.

In this paper, we propose BeamsNet,\hlc{an end-to-end deep learning approach aiming to replace the, commonly used, model-based approach for estimating the AUV velocity vector based on the DVL beam measurements.} \hlcc{To that end, we leverage from the well-known deep learning capabilities such as noise reduction, the ability to capture non-linearity behavior, and other uncertainty properties in the data.}
Two versions of BeamsNet, differing in their input to the network, are suggested. The first uses the current DVL beam measurements and inertial sensors (accelerometers and gyroscopes) data, while the other utilizes only DVL data taking the current and past DVL measurements for the regression process. We did a simulation and sea experiments to validate the proposed learning approach compared to the model-based approach. The latter experiments took place in the Mediterranean Sea using the University of Haifa's Snapir AUV. Approximately four hours of recorded data were collected, including the accelerometers, gyroscopes, and DVL measurements. 
\hlcc{Generally, In most AUVs the DVL is used for determining the platform's position in a dead reckoning approach, which means that by integrating the DVL estimated velocity vector over time, the position solution will be provided. Therefore, by significantly improving the DVL estimated velocity  accuracy, in turn, the position accuracy will be improved. The accuracy of the AUV's position is crucial due to the nature of the missions and the need to navigate autonomously.\\
To summarize, this paper's contributions are as follows:}
\begin{enumerate}
    \item \hlcc{BeamsNet, a deep learning framework suggested to replace the model-based approach using the same input.}
    \item \hlcc{An additional BeamsNet architecture that leverages from  inertial sensor readings.}
    \item \hlcc{A} \href{https://github.com/ansfl/BeamsNet}{GitHub repository} \hlcc{containing our code and dataset as a benchmark and to encourage further research in the field.}   
\end{enumerate}

The rest of the paper is organized as follows: Section \ref{DVL} describes the DVL equations and error models. Section \ref{data} introduces the proposed approach and the network architecture. In Section \ref{AandR} the results from the simulation and the sea experiment are presented, and in Section \ref{con} the conclusions are discussed. 
\section{DVL Velocity Calculations} \label{DVL}
\begin{figure}[h]
	\centering
		\includegraphics[scale=.30,]{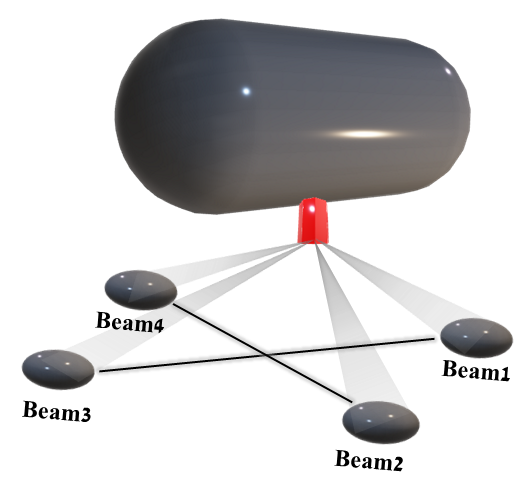}
	  \caption{DVL's transducers configured in a '$x$' shape, also known as a "Janus Doppler configuration".}\label{fig1}
\end{figure}

\begin{figure}[h!]
	\centering
		\includegraphics[scale=0.35]{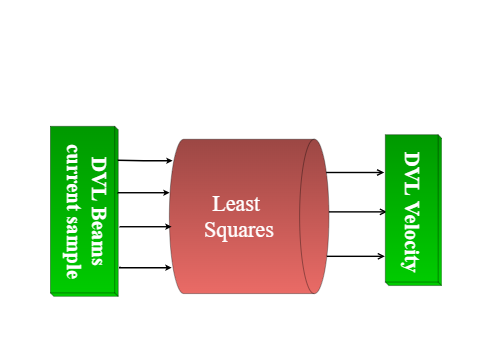}
	  \caption{An illustration of the Least Squares operation. The input is the beam measurements, and the output is the estimated DVL velocity vector.}\label{fig2}
\end{figure}

The DVL sensor operates by transmitting acoustic beams in four directions and receiving the deflected signals from the seafloor. Based on the Doppler effect and the frequency shift between the transmitted and the received signals, the DVL can determine the AUV velocity. The DVL is both the transmitter and the receiver. Therefore intermediate reflection of the acoustic beam at the ocean floor is treated as a stationary receiver immediately followed by a stationary transmitter. Taken from \cite{brokloff1994matrix}, the expressions \eqref{eqn:1}-\eqref{eqn:3} show the relation between the frequency and the beam velocity. The frequency measured in the receiver is
\begin{equation}\label{eqn:1}
    \centering
        f_{r}=f_{t}\left(\frac{1\mp\frac{\upsilon_{beam}}{c}}{1\pm\frac{\upsilon_{beam}}{c}}\right)
\end{equation} 
Where $f_{r}$ and $f_{t}$ are the received and the transmitted frequency, respectively, $\upsilon_{beam}$ is the beam velocity, and $c$ is the speed of sound. By multiplying the conjugate of the denominator and assuming the speed of the vehicle is less than the speed of sound, the squared terms can be ignored. Therefore, the frequency shift \hlc{$\Delta f$,} is approximately
\begin{equation}\label{eqn:2}
    \centering
        \Delta f\approx \frac{2f_{t}\upsilon_{beam}}{c}
\end{equation} 
The DVL's transducers are commonly configured in a '$\times$' shape configuration, known in the literature as the "Janus Doppler configuration", as seen in Figure \ref{fig1}.
\newline
The beam velocity in each direction can be defined as: 
\begin{equation}\label{eqn:3}
    \centering
        \boldsymbol{\upsilon}_{beam}=\frac{c}{2f_{t}}\Delta\boldsymbol{f}
\end{equation} 
By observing the geometric relationship between the DVL beams and the DVL body, the direction of each beam in the DVL's body frame can be expressed as \cite{liu2018ins}: 
\begin{equation}\label{eqn:4}
    \centering
        \boldsymbol{b}_{\dot{\imath}}=
        \begin{bmatrix} 
        \cos{\psi_{\dot{\imath}}}\sin{\alpha}\quad
        \sin{\psi_{\dot{\imath}}}\sin{\alpha}\quad
        \cos{\alpha}
    \end{bmatrix}_{1\times3}
\end{equation} 
where $\dot{\imath}=1,2,3,4$ represents the beam number and $\psi$ and $\alpha$ are the yaw and pitch angles relative to the body frame, respectively \cite{tal2017inertial}. The pitch angle is fixed and has the same value for each beam, and the yaw angle can be expressed by \cite{yona2021compensating}:
\begin{equation}\label{eqn:5}
    \centering
        \psi_{\dot{\imath}}=(\dot{\imath}-1)\cdot\frac{\pi}{2}+\frac{\pi}{4}\;[rad]\;,\; \dot{\imath}=1,2,3,4
\end{equation} Thus, by defining a transformation matrix $\mathbf{H}$, the relation between the DVL velocity in body frame, $\boldsymbol{v}_{b}^{b}$, to the beam velocity measurements, $\boldsymbol{\upsilon}_{beam}$, can be written as follows:
\begin{equation}\label{eqn:6}
    \centering
        \boldsymbol{\upsilon}_{beam}=\mathbf{H}\boldsymbol{v}_{b}^{b} ,\quad
        \mathbf{H}=
        \begin{bmatrix} 
            \boldsymbol{b}_{1}\\\boldsymbol{b}_{2}\\\boldsymbol{b}_{3}\\\boldsymbol{b}_{4}\\
    \end{bmatrix}_{4\times3}
\end{equation} 
To model the measured beam velocities, a beam error model is required. To that end, a bias \hlcc{, scale factor,} and a zero white mean Gaussian noise are added to the beam velocity measurements from \eqref{eqn:6}, yielding 
\begin{equation}\label{eqn:7}
    \centering
        \boldsymbol{y}= \mathbf{H}(\boldsymbol{v}_{b}^{b}\cdot(1+s_{DVL})+\boldsymbol{b}_{DVL}+\boldsymbol{n}
\end{equation}
where $\boldsymbol{b}_{DVL}$ is the bias $4\times1$ vector with different values, $s_{DVL}$ \hlcc{is the scale factor} $4\times1$ vector with different values, $\boldsymbol{n}$ is the zero white mean Gaussian noise\hlcc{, and }$\boldsymbol{y}$ is the beam velocity measurements.\\
Once the beam velocity measurements are obtained, the DVL velocity needs to be estimated. To that end, a Least Squares (LS) estimator is used:
\begin{equation}\label{eqn:8}
    \centering
        \hat{\boldsymbol{v}}_{b}^{b}=
        \underset{\boldsymbol{v}_{b}^{b}}{\argmin}{\mid\mid\boldsymbol{y}-\mathbf{H}\boldsymbol{v}_{b}^{b} \mid\mid}^{2}
\end{equation} 
The solution for this estimator \hlc{$\hat{\boldsymbol{v}}_{b}^{b}$, the DVL velocity vector}, is the pseudo inverse of matrix $\mathbf{H}$ times the beams velocity measurement $\boldsymbol{y}$, as can be seen below \cite{braginsky2020correction}:
\begin{equation}\label{eqn:9}
    \centering
        \hat{\boldsymbol{v}}_{b}^{b}=(\mathbf{H}^{T}\mathbf{H})^{-1}\mathbf{H}^{T}\boldsymbol{y}
\end{equation} 
The solution to the LS estimator does two operations. The first is filtering the bias and the noise, and the second is transforming the beam velocity measurements to the DVL velocity. The LS operation is illustrated in Figure \ref{fig2}. 
\section{data-driven DVL Velocity Calculations}\label{data}
As mentioned above, to determine the DVL's velocity from the beam velocity measurements, a LS estimator \eqref{eqn:9} is used. \\
Besides its simplicity, the LS estimator's advantage is that when assuming independent observations normally distributed with a constant variance, the LS estimator coincides with the maximum-likelihood estimator. In this case, the LS estimator is considered optimal among all the unbiased estimators because it achieves the Cramer-Rao Bound (CRB) \cite{stoica1989music}. \\
One disadvantage of the LS estimator is its sensitivity to outliers \cite{sohn1997detection}. Also, when the noise is not Gaussian, it is unnecessarily the optimal estimator, and its performance may be questionable \cite{myung2003tutorial}; \cite{bar2004estimation}. \\
To overcome those disadvantages, in this paper, we drive an end-to-end data-driven approach to estimate the DVL velocity vector instead of using the LS approach. \\
Our motivation stems from the literature showing that data-driven approaches were shown to create features enabling noise reduction in several different domains. For example, in the inertial sensor field, \cite{brossard2020denoising} based their work on convolutional neural network and feature selection and were able to denoise gyroscope data from low-cost IMU to receive accurate attitude estimates. Furthermore, hybrid deep recurrent neural networks showed good results in the field of low-cost IMU noise reduction \cite{han2021hybrid}.     \\
Two different data-driven approaches to regress the DVL velocity vector are suggested:
\begin{itemize}
    \item \textbf{IMU/DVL based BeamsNet (BeamsNetV1)}: Employs current beam measurements and IMU readings to regress the current DVL velocity vector using 1DCNN.
    \item \textbf{DVL based BeamsNet (BeamsNetV2)}: Uses $n$ past DVL measurements and the current beam measurements to regress the current DVL velocity vector using 1DCNN.
\end{itemize}
In the following subsections, we elaborate on the two architectures. 
\subsection{IMU/DVL based BeamsNet (BeamsNetV1)}
\begin{figure}[h!]
	\centering
		\includegraphics[scale=0.28]{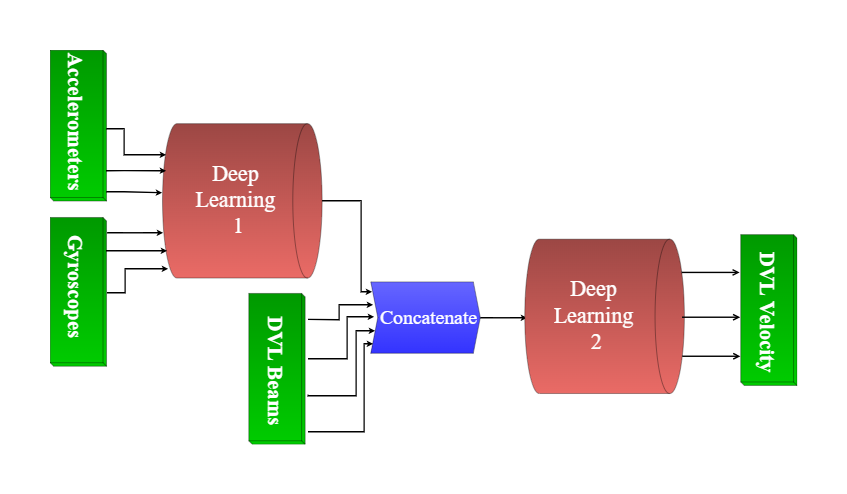}
	  \caption{BeamsNetV1 multi-head structure consists of three heads, one operating on accelerometer data, the second one on the gyroscope data, and the third on the current DVL beam measurements.}\label{fig3}
\end{figure}

In this approach, we propose to use current DVL beam measurements and IMU readings to regress the current DVL velocity vector, as illustrated in Figure \ref{fig3}. 

\begin{figure}[h!]
	\centering
		\includegraphics[scale=0.32]{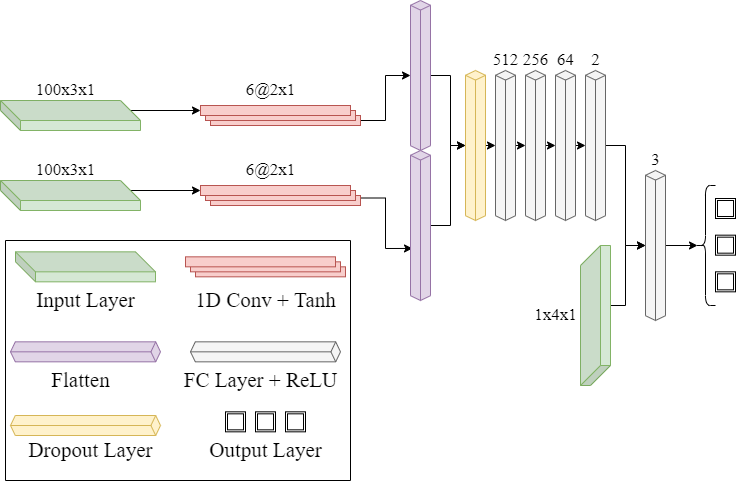}
	  \caption{BeamsNetV1 network architecture.}\label{fig4}
\end{figure}

Thus, the input to the network is the accelerometers, gyroscopes, and DVL raw measurements, and the output is the estimated DVL velocity vector. 
As the IMU's inertial sensors capture the vehicle dynamics, its measurements may improve the DVL beam measurements while estimating the velocity vector. Consequently, to apply this approach, an IMU is needed where in the model-based velocity estimation, \eqref{eqn:9}, only a DVL is required. In addition, the regressed DVL velocity vector is now a function of both IMU and DVL measurements. Hence, when used as an update within a nonlinear filter, one must take into account this process (IMU) and measurement (DVL) cross-covariance correlation \cite{simon2006optimal,klein2021ins,eliav2018ins}; 
On the other hand, only the current DVL beam measurements are used, removing the dependence of the dynamics of the AUV  (if past DVL measurements are also utilized). \\
The DVL low-rate sensors have typical values of $1$Hz, while the inertial sensors provide their measurements at a high rate with typical values of $100$Hz. Thus, until the current DVL measurement is available, one hundred measurements are available from each of the inertial sensors (three accelerometers and three gyroscopes). To cope with the different input sizes, BeamsNetV1 architecture contains three heads. The first is for the $100$ samples of the three-axes accelerometer, and the second is for the $100$ samples of the three-axes gyroscope, operating simultaneously. The last head takes the DVL beam measurements.\\
The raw accelerometer and gyroscopes measurements pass through a one-dimensional convolutional (1DCNN) layer consisting of six filters of size $2\times 1$ that extract features from the data. Next, the features extracted from the accelerometers and gyroscopes are flattened, combined, and then passed through a dropout layer with $p=0.2$. After a sequence of fully connected layers, the current DVL measurement is combined and goes through the last fully connected layer that produces the $3\times 1$ vector, which is the estimated DVL velocity vector. The architecture and the activation functions after each layer are presented in Figure \ref{fig4}.
\subsection{DVL based BeamsNet (BeamsNetV2)}
\begin{figure}[h!]
	\centering
		\includegraphics[scale=0.22]{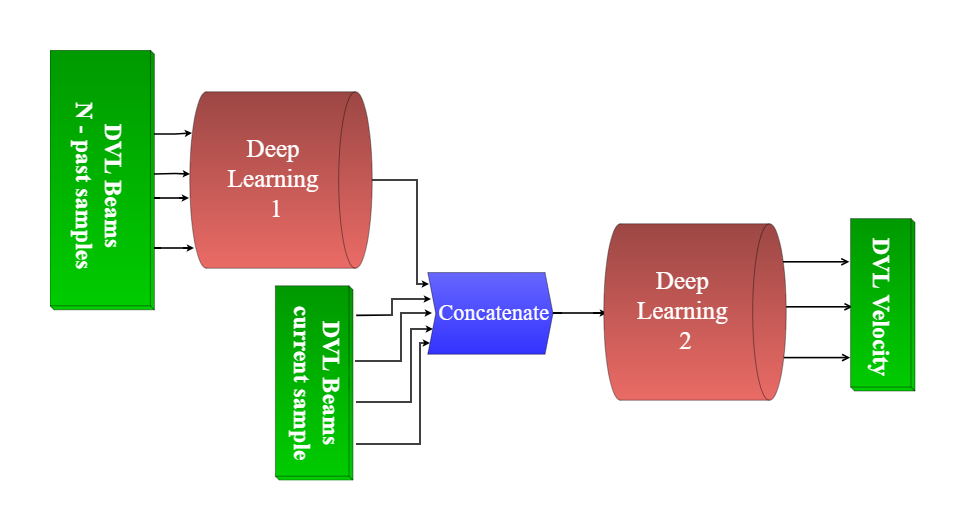}
	  \caption{BeamsNetV2 multi-head structure with $n$ past beam measurements,  used as input to the first head and the second head receives the current DVL beam measurements to regress the DVL velocity vector.}\label{fig5}
\end{figure}

\begin{figure}[h!]
	\centering
		\includegraphics[scale=0.32]{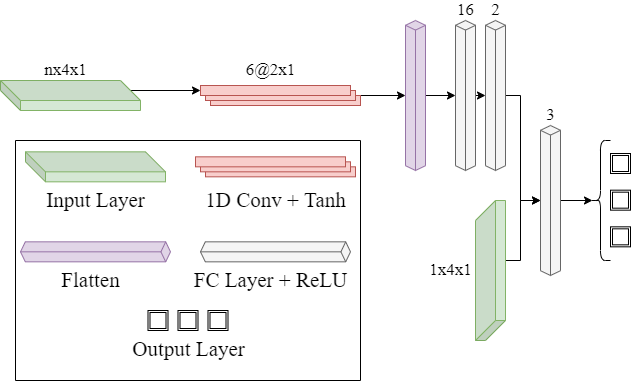}
	  \caption{BeamsNetV2 network structure. }\label{fig6}
\end{figure}

In this approach, we propose to use past DVL beam measurements in addition to the current beam measurements to estimate the velocity vector of the DVL. To apply this approach, no IMU is needed, only several past DVL beam measurements. The underlying assumption is that the AUV has low maneuvering capabilities and, most of the time, travels in straight-line trajectories. Therefore, past measurements are likely to contain the same AUV dynamics as the current one and hence may improve the estimation of the AUV velocity vector. The number of past measurements to use is treated as a hyper-parameter. The proposed approach is shown in Figure \ref{fig5}. 
When only the DVL data is available, a two-headed 1DCNN can be used and presented in Figure \ref{fig6}. The network's input is $n$ past samples of the DVL beam measurements. Same as for the BeamsNetV1 architecture, the input goes through a one-dimensional convolutional layer with the same specifications. The output from the convolutional layer is flattened and passes through two fully connected layers. After that, the current DVL measurement is combined with the last fully connected layer output and goes into the last fully connected layer that generates the output. In this paper, we used $n=3$ past measurements on the DVL.
\subsection{BeamsNet Training Process}
The objective of the training is to determine the weights and biases that solve the given problem. Fully connected layers are built by a number of neurons. The computation performed by each neuron is
\begin{equation}\label{eqn:10}
    \centering
        z_{\dot{\imath}}^{(\ell)}=\sum_{\dot{\jmath}=1}^{n_{\ell-1}}\boldsymbol{\omega}_{\dot{\imath}\dot{\jmath}}^{(\ell)}\boldsymbol{a}_{\dot{\jmath}}^{(\ell-1)}+\boldsymbol{b}_{\dot{\imath}}^{(\ell)}
\end{equation} where
\begin{description}[font=$\bullet$~\normalfont\scshape]
\item $\boldsymbol{\omega}_{\dot{\imath}\dot{\jmath}}^{(\ell)}$ is the weight of the $\dot{\imath}^{th}$ neuron in the $\ell^{th}$ layer associated with the output of the $\dot{\jmath}^{th}$ neuron in the $(\ell-1)^{th}$ layer.
\item $\boldsymbol{b}_{\dot{\imath}}^{(\ell)}$ represents the bias in layer $\ell$ of the $\dot{\imath}^{th}$ neuron.
\item $\boldsymbol{n}_{\ell-1}$ represents the number of neurons in the $\ell-1$ layer.
\end{description}
The output of the neuron $ z_{\dot{\imath}}^{(\ell)}$ is passed through a nonlinear activation function, $h(\cdot)$ and the output is defined as
\begin{equation}\label{eqn:11}
    \centering
        a_{\dot{\imath}}^{(\ell)}=h(z_{\dot{\imath}}^{(\ell)}).
\end{equation}
The behavior characterization of a single neuron is expressed by these two equations \cite{gonzalez2018deep}. As for the convolutional layer, we assume a $m_{1} \times m_{2}$ filter (or kernel) and the output of the layer can be written as follows:
\begin{equation}\label{eqn:12}
    \centering
        C_{\dot{\imath}\dot{\jmath}}^{(\ell)}=\sum_{\alpha=0}^{m_{1}}\sum_{\beta=0}^{m_{2}}\boldsymbol{\omega}_{\alpha\beta}^{(r)}\boldsymbol{a}_{(\dot{\imath}+\alpha)(\dot{\jmath}+\beta)}^{(\ell-1)}+\boldsymbol{b}^{(r)}
\end{equation} where
\begin{description}[font=$\bullet$~\normalfont\scshape]
\item $\boldsymbol{\omega}_{\alpha\beta}^{(r)}$ is the weight in the $(\alpha,\beta)$ position of the $r^{th}$ convolutional layer.
\item $\boldsymbol{b}^{(r)}$ represents the bias of the $r^{th}$ convolutional layer.
\item $\boldsymbol{a}_{\dot{\imath}\dot{\jmath}}^{(\ell-1)}$ is the output of the preceding layer.
\end{description}
Two different activation functions are used in the network structure
\begin{enumerate}
    \item \textbf{Rectified Linear Unit (ReLU)} is an activation function with a strong mathematical and biological basis. It took a big role in improving the training of deep neural networks \cite{agarap2018deep}.\\ 
    The ReLU activation function is defined by 
    \begin{equation}\label{eqn:13a}
    \centering
        ReLU(z_{\dot{\imath}}^{(\ell)})=max(0,z_{\dot{\imath}}^{(\ell)}).
   \end{equation}
    \item \textbf{Hyperbolic Tangent Function (Tanh)} is a continuous and differentiable function, and the values are bounded between -1 and 1. Different signs of outputs from previous layers are used as inputs in the next layer \cite{sharma2017activation}.\\ 
    The Tanh activation function is defined by 
    \begin{equation}\label{eqn:13b}
    \centering
     Tanh(z_{\dot{\imath}}^{(\ell)})=\frac{e^{z_{\dot{\imath}}^{(\ell)}}-e^{-z_{\dot{\imath}}^{(\ell)}}}{e^{z_{\dot{\imath}}^{(\ell)}}+e^{-z_{\dot{\imath}}^{(\ell)}}}.
    \end{equation}
\end{enumerate}
The mean squared error (MSE) loss function is employed for the regression process
\begin{equation}\label{eqn:14}
    \centering
        J(\boldsymbol{y}_{\dot\imath},\hat{\boldsymbol{y}}_{\dot\imath})=\frac{1}{n}\mid\mid \boldsymbol{y}_{\dot\imath}-\hat{\boldsymbol{y}}_{\dot\imath}\mid\mid^{2}
\end{equation} where $\boldsymbol{y}_{\dot\imath}$ is the ground truth and $\hat{\boldsymbol{y}}_{\dot\imath}$ is the predicted value. The process of data going through equations \eqref{eqn:10} - \eqref{eqn:14} is known as the forward propagation that generates the prediction $\hat{\boldsymbol{y}}_{\dot\imath}$ \cite{zhao2017convolutional}. As a means to update the weights and biases, a gradient decent approach is implemented 
\begin{equation}\label{eqn:15}
    \centering
        \boldsymbol{\theta}=\boldsymbol{\theta}-\eta\nabla_{\theta}J({\boldsymbol{\theta}})\; , \quad  \boldsymbol{\theta}=[\omega\quad b]^{T}
\end{equation}
where 
\begin{description}[font=$\bullet$~\normalfont\scshape]
\item $\boldsymbol{\theta}$ is the vector of weights and biases.   
\item $\eta$ is the learning rate.
\item $J(\boldsymbol{\theta})$ is the loss function with respect to the vector $\boldsymbol{\theta}$.
\item \hlc{$\nabla_{\theta}$ is the Gradient operator}.
\end{description}
To that end, an adaptive learning rate method, RMSprop, is applied as it aims to resolve the radically diminishing learning rates. The RMSprop divides the learning rate by an exponentially decaying average of squared gradients \cite{ruder2016overview}.

\section{Analysis and Results}\label{AandR}
This section presents simulation and sea experiment results. \hlcc{Several matrices, commonly used for performance assessment purposes of the AI techniques}\cite{armaghani2021comparative},  \hlcc{were chosen for evaluating the suggested framework: 1) root mean squared error (RMSE) 2) mean absolute error (MAE) 3) the coefficient of determination ($R^{2}$), and 4) the variance account for (VAF). The RMSE and MAE express the velocity error in units of $[m/s]$, while the $R^{2}$ and VAF are unitless.  Those matrices are defined as follows:}
\begin{equation}\label{eqn:16}
    \centering
        RMSE(\boldsymbol{x}_{\dot\imath},\hat{\boldsymbol{x}}_{\dot\imath})=\sqrt{\frac{\sum_{\dot\imath=1}^{N}(\boldsymbol{x}_{\dot\imath}-\hat{\boldsymbol{x}}_{\dot\imath})^{2}}{N}}
\end{equation}
\begin{equation}\label{eqn:17}
    \centering
        MAE(\boldsymbol{x}_{\dot\imath},\hat{\boldsymbol{x}}_{\dot\imath})=\frac{\sum_{\dot\imath=1}^{N}|\boldsymbol{x}_{\dot\imath}-\hat{\boldsymbol{x}}_{\dot\imath}|}{N}
\end{equation}
\begin{equation}\label{eqn:18}
    \centering
        R^{2}(\boldsymbol{x}_{\dot\imath},\hat{\boldsymbol{x}}_{\dot\imath})=1- \frac{\sum_{\dot\imath=1}^{N}(\boldsymbol{x}_{\dot\imath}-\hat{\boldsymbol{x}}_{\dot\imath})^{2}}{\sum_{\dot\imath=1}^{N}(\boldsymbol{x}_{\dot\imath}-\bar{\boldsymbol{x}}_{\dot\imath})^{2}}
\end{equation}
\begin{equation}\label{eqn:19}
    \centering
        VAF(\boldsymbol{x}_{\dot\imath},\hat{\boldsymbol{x}}_{\dot\imath})=[1-\frac{var(\boldsymbol{x}_{\dot\imath}-\hat{\boldsymbol{x}}_{\dot\imath})}{var(\boldsymbol{x}_{\dot\imath})}]\times100
\end{equation}
where N is the number of samples, $\boldsymbol{x}_{\dot\imath}$ is the ground truth velocity vector norm of the DVL,\hlc{ $\hat{\boldsymbol{x}}_{\dot\imath}$ is the predicted velocity vector norm of the DVL generated by the network, $\bar{\boldsymbol{x}}_{\dot\imath}$ is the mean of the ground truth velocity vector norm of the DVL, and $var$ stands for variance. Note that if the VAF is $100$, the $R^{2}$ is $1$, and the RMSE and MAE are $0$, the model will be considered outstanding. }
\subsection{Simulation}\label{Simulation}
To examine the proposed approach, a straight line trajectory of an AUV was simulated in three different constant speeds: $1[m/s], 2[m/s]$, and $3[m/s]$. Then, the actual DVL beam measurements were calculated. To create the measured beam velocities, a DVL beam model in \eqref{eqn:7} \hlcc{was employed}. 

\begin{figure}[h!]
    \centering
      {\includegraphics[scale=0.5]{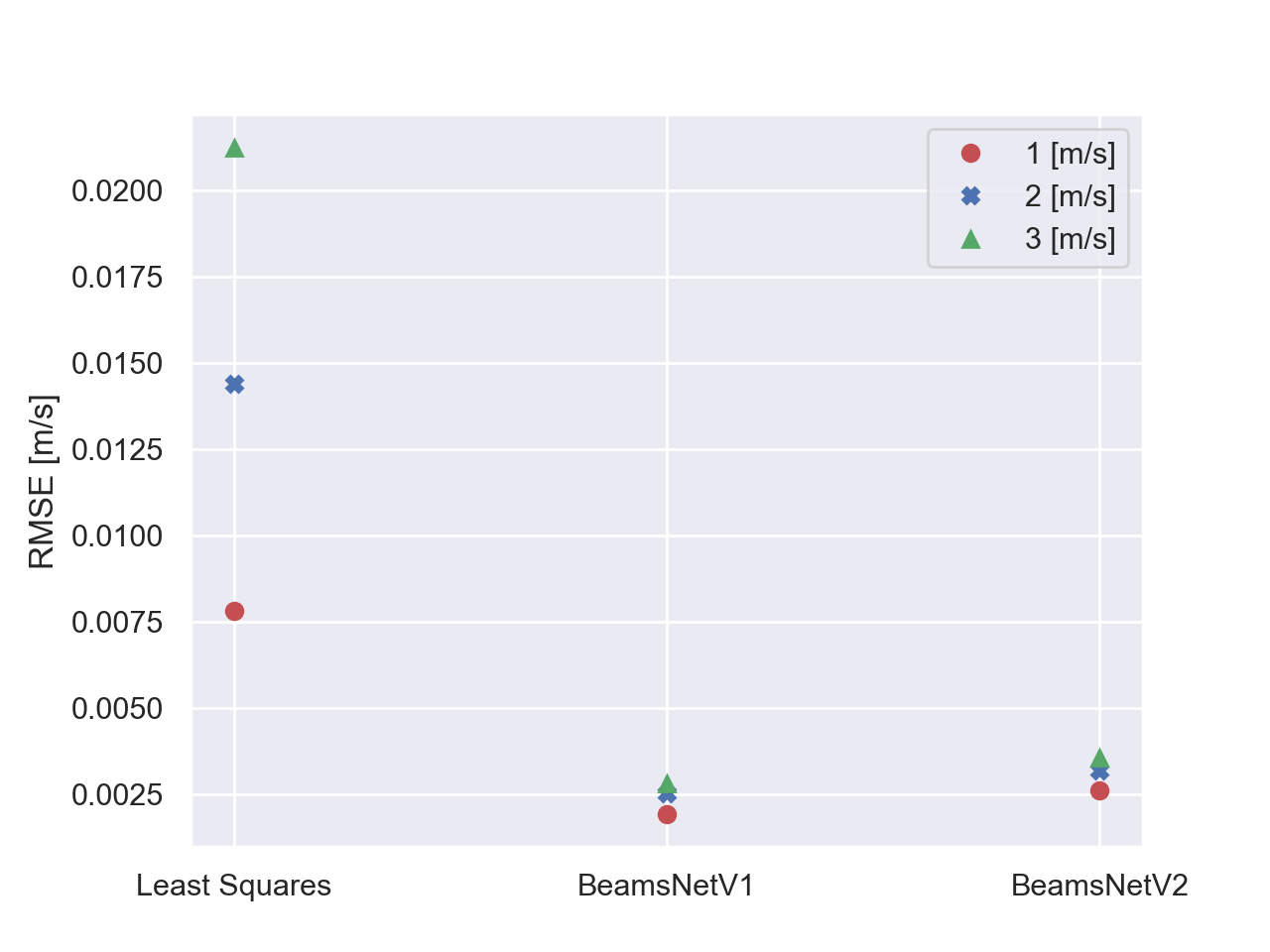}}
  	  \caption{The performance of each method with respect to the RMSE for different constant velocities of the AUV. }\label{fig7}
\end{figure}

%
For the analysis, the scale factor is set to $0.7 \%$,
the bias is $0.0001[m/s]$, and the white zero-mean Gaussian noise standard deviation is $0.042[m/s]$. The time duration of each trajectory is 120 minutes, corresponding to $7,200$ DVL measurements (1Hz sampling rate) and $720,000$ IMU samples (100Hz sampling rate). Those measurements were divided into a $75\%$ train set and $25\%$ test set without shuffling the data. The simulated data was tested on BeamsNetV1, BeamsNetV2, and the LS approach. The learning rate was set to $\eta=0.01$ with a learning rate decay of 0.1 every 15 epochs. The data was divided into batches of size 4 and trained over 30 epochs. \\
\hlcc{To evaluate the simulation results, only the RMSE metric was used. Figure}~\ref{fig7} \hlcc{presents the RMSE of the LS approach, as well as the suggested networks}. There is an indication that both BeamsNetV1 and BeamsNetV2 improve the DVL velocity vector estimation significantly when compared to the commonly used LS approach. The LS method produces a bigger RMSE for higher AUV velocities, whereas the suggested methods decrease the RMSE significantly and the difference between different DVL velocities is relatively small.   
\\
To emphasize BeamsNet performance, their RMSE improvement is presented in Table \ref{table:1}. 
 \begin{table}[h!]
        \centering
            \caption{The improvement of the RMSE velocity measurements using BeamsNetV1 and BeamsNetV2 when compared to the known LS approach in [\%].}
            \begin{tabular}{ |c|c|c|c| }
            \hline
            Velocity $[m/s]$ & BeamsNetV1 $[\%]$ & BeamsNetV2 $[\%]$\\
            \hline
                1 & $75.45$ & $66.72$  \\ 
            \hline
                2 & $82.451$ & $77.91$  \\ 
            \hline
                3 & $86.76$ & $83.32$ \\ 
            \hline
            \end{tabular}
\label{table:1}
\end{table}\\
The table shows the percentage of improvement BeamsNetV1 and BeamsNetV2 provide with respect to the RMSE of the estimated DVL velocity when compared to the RMSE of the LS approach. The results indicate that the faster the AUV travels, the better the improvement, and the IMU data's effect helps improve the estimated DVL velocity measurements even further.  

\subsection{AUV Sea Experiments}
\begin{figure}[h!]
	\centering
		\includegraphics[scale=0.35]{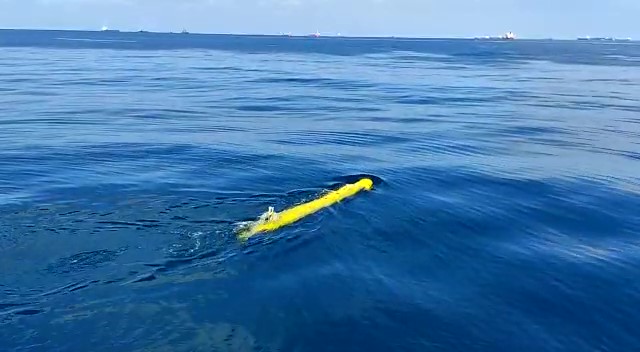}
	  \caption{The Snapir AUV during a mission in the Mediterranean Sea.}\label{fig8}
\end{figure}

To validate the proposed approach and simulation results, sea experiments were conducted. They took place in the Mediterranean Sea using the "Snapir" AUV \url{https://www.marinetech.haifa.ac.il/ocean-instruments}.
The Snapir is an A18-D, ECA GROUP mid-size AUV for deep water applications. Capable of rapidly and accurately mapping large areas of the sea floor, Snapir has a length of $5.5 [m]$, a diameter of $0.5 [m]$, 24 hours' endurance, and a depth rating of $3000 [m]$. Snapir carries several sensors as its payload, including an interferometric authorized  synthetic aperture sonar (SAS) and Teledyne RD Instruments, Navigator DVL \cite{Teledyne}. 
Figure \ref{fig8} shows the Snapir AUV during a mission.
\newline
The dataset was created by collecting DVL data from nine different missions performed by the AUV with a total time duration of 13,886 seconds, which translates to the same number of DVL measurements and 1,388,600 IMU measurements.  
This dataset is described in \cite{shurin2022autonomous} and can be found on \url{ https://github.com/ansfl/Navigation-Data-Project/}. 
Each of the missions had different parameters regarding the length of the mission, the objective, the speed of the AUV, the depth of the AUV, and the maneuvers it performed. In ideal circumstances, two DVLs would be located in the AUV in order to use one as ground truth and the second  as the unit under test. Since this was not the case, the estimated DVL velocity given by the DVL was placed in the DVL beam velocity error model \eqref{eqn:17} in the same manner as was used in the simulation (see section \ref{Simulation}). The scale factor, bias, and STD of the zero-mean white Gaussian noise were 0.7\%, 0.0001[m/s], and 0.042[m/s] respectively. Thus the measurements were considered as if they were taken from the DVL under test, while the readings from the experiment were considered  the ground truth. Furthermore, we examined a different approach that adds zero-mean white Gaussian noise with STD of 0.0001[m/s] to the experiment DVL data and then passes it through the DVL beam velocity error model. Both approaches had similar results, and therefore we choose to discuss only the former.\\ 

\begin{figure}[h!]
    \centering
      {\includegraphics[scale=0.5]{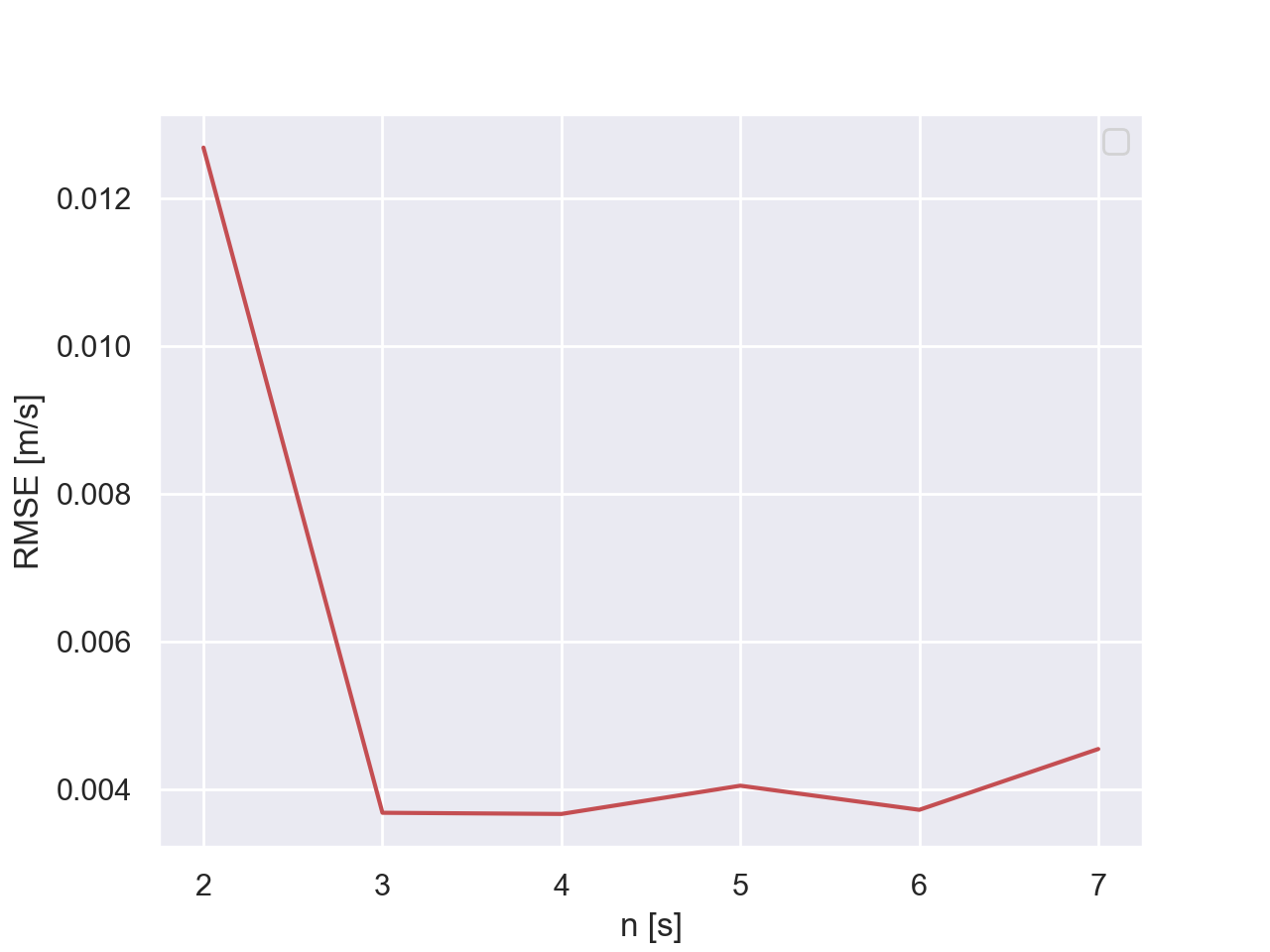}}
  	  \caption{The RMSE of the DVL estimated velocity vector as a function of $n$ past DVL beam measurements. }\label{fig9}
\end{figure}

The dataset was divided into $75\%$ train set and $25\%$ test set without shuffling the data. The data was used to train and test BeamsNetV1 and BeamsNetV2. A learning rate of $\eta=0.001$ with a learning rate decay of 0.1 every 15 epochs was implemented. The data was divided into batches of size 4 and trained over 50 epochs. \\
First, the number of past beam measurements to use was determined. To that end, we examined a range between two to seven past beam measurements with a corresponding duration of one to seven seconds. The estimated DVL velocity RMSE as a function of the number of past measurements is given in Figure \ref{fig9}.

By looking at different values of $n$, a number of past samples of the DVL, it was revealed that an optimal number provides the best solution. The highest number that was tested is seven because it takes seven seconds to obtain this data (due to the $1$ Hz sampling rate of the DVL), which requires a bigger storage unit, which, it seems, is not required. In BeamsNetV2, $n=3$ past measurements were used because, as Figure \ref{fig9} suggests, it provides the smallest RMSE.

Next, the performances of the two BeamsNet approaches were compared to the LS approach using the test dataset. Those results are summarized in Table \ref{table:3} and show that the suggested methods improve the estimated DVL velocity when compared to the known LS approach. 
Both methods, BeamsNetV1 and BeamsNetV2, showed an improvement of 64.75\% and 62.86\%, respectively, when looking at the RMSE of the estimated DVL velocity norm.
\hlcc{The MAE metric indicates that BeamsNetV1 preform better using the additional inertial sensor data when comparing to BeamsNetV2 and LS that uses only the DVL data. Finally, the $R^{2}$ and VAF matrices show that all the approaches have good statistical performance.}\\
\begin{table*}[h!]
        \centering
            \caption{Estimated DVL velocity norm RMSE, MAE, $R^{2}$ and VAF of BeamsNet and the improvement compared to the LS approach.}
            \begin{tabular}{ |c|c|c|c|c|c|c| }
            \hline
             Method & RMSE $[m/s]$ & MAE $[m/s]$ & $R^{2}$ & VAF & RMSE Improvement [\%]\\
            \hline
            BeamsNetV1 (ours) & $0.003503$ & $0.002817$ & $0.999971$ & $99.997128$ & $64.75$ \\ 
            \hline
            BeamsNetV2 (ours) & $0.003690$ & $0.004365$ & $0.999738$ & $99.973886$ & $62.86$  \\ 
            \hline
            LS (baseline) & $0.009938$ & $0.008445$ & $0.999773$ & $99.991989$ & N/A \\
            \hline
            \end{tabular}
\label{table:3}
\end{table*}

\subsection{Discussion on the Network Structure and Convergence}
\begin{figure}
\begin{subfigure}{\columnwidth}
    \centering
    \includegraphics[width=\textwidth]{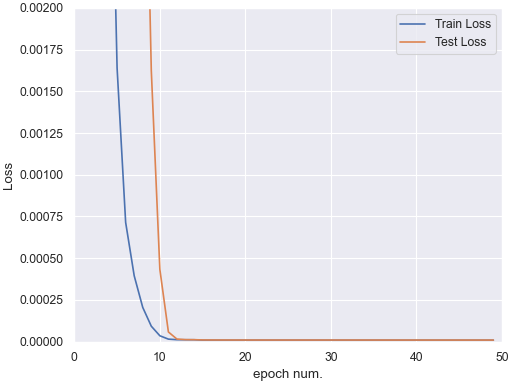}  
    \caption{BeamsNetV1}
    \label{SUBFIGURE LABEL 1}
\end{subfigure}
\begin{subfigure}{\columnwidth}
    \centering
    \includegraphics[width=\textwidth]{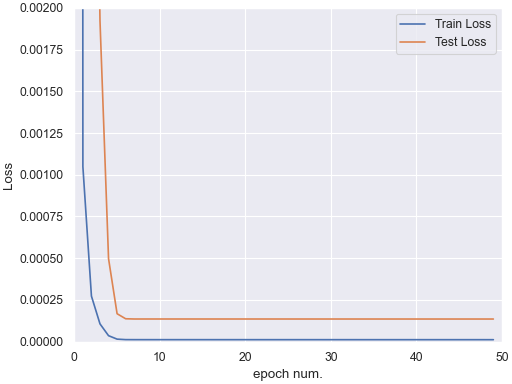}  
    \caption{BeamsNetV2}
    \label{SUBFIGURE LABEL 2}
\end{subfigure}
\caption{Loss values as a function of the epoch number for (a) BeamsNetV1 and (b) BeamsNetV2}
\label{fig10}
\end{figure}

While experimenting with the data and the deep learning framework, several insights regarding the architecture and the hyper-parameters were observed. First, the head containing the current DVL beam measurements in the networks should be placed deep in the network, preferably before the output layer. Placing it in the first layers degrades the network performance. Additionally, the data was divided into mini-batches: a batch size of four significantly improved the network accuracy, while bigger batch sizes damaged the network performance. These two attributes were the key changes that made the networks perform well \hlcc{and converge }.\hlcc{Besides the hyper-parameters, and since most of the activation functions in the suggested network are ReLUs, the weights were randomly initialized with the Kaiming uniform method, designed to improve performance for non-symmetrical activation functions} \cite{he2015delving}.\hlcc{ To examine the convergence of the network, the loss function values were examined as a function of the epoch number for both the training and test data and can be seen in figure }\ref{fig10}. \hlcc{The graphs indicates that the training minimizes the loss function and that there is no overfitting. }

\section{Conclusions}\label{con}
In this paper,\hlc{an end-to-end deep learning approach aiming to replace the LS filter (the commonly used model-based approach) was proposed for estimating the AUV velocity vector based on the DVL beam measurements.} To that end, two approaches were derived: BeamsNetV1 used current DVL beam measurements and inertial data; and BeamsNetV2 utilized only the DVL beam measurements, taking the current and $n$ past measurements. \\
First, a dataset was generated in a simulation to evaluate the proposed approach. Different constant speeds on a straight-line trajectory were simulated with proper sensor readings. The simulation showed that the suggested approaches give better results than the model-based approach, showing an improvement of $66\%-87\%$. In addition, it was observed  that the faster the AUV travels, the better both BeamsNet approaches perform. \\
To further evaluate the proposed approaches, sea experiments were conducted in the Mediterranean Sea, using the University of Haifa's Snapir AUV. Data from different missions containing several different trajectories and velocities was combined together, resulting in four hours of recorded DVL and inertial data. The recorded data was used to train and test BeamsNetV1 and BeamsNetV2, and the results showed a significant improvement compared to the LS method with respect to the RMSE metric. Specifically, an improvement of $64.75\%$ in BeamsNetV1 when both DVL and the inertial sensors are available and a $62.86\%$ improvement when using three past DVL measurements in BeamsNetV2. \hlc{In addition, the MAE criteria suggested that the additional information from the IMU readings in BeamsNetV1, results in better estimation when comparing to BeamsNetV2 and LS that employ only DVL data. The statistical performance criteria $R^{2}$ and VAF, indicates that all the discussed approaches are capable of dealing with the estimation problem.} \\
As both BeamsNet approaches obtained similar performance, it is important to emphasize the pros and cons of each architecture. As BeamsNetV1 requires both inertial and DVL measurements, the regressed DVL velocity
vector is now a function of both IMU and DVL. Hence, when used as updated within a nonlinear filter, one must take into account this process (IMU) and measurement (DVL) cross-covariance correlation. Yet, in this architecture,  only the current DVL beam measurements are used,
removing the dependence on the dynamics of the AUV (if
past DVL measurements were also utilized). In BeamsNetV2, only current and past DVL measurements are used, neglecting the usage of an IMU. However, if the AUV changes its dynamics during the time period in which the past measurements are taken, the performance of the approach may degrade. \\
\hlcc{To conclude, DVL is commonly used in AUVs for position determination in a dead reckoning approach. Hence, improving the estimated AUV velocity accuracy will result in a more accurate position vector. To successfully complete its task, the AUV position accuracy is critical, and this paper offers a method for a more precise position solution. In some AUVs, the IMU sensor is used only for attitude determination. Now, using the proposed approach, IMU measurements can also be utilized to improve the AUV velocity estimation. In addition, the improved performance may allow AUV designers to select a lower grade DVL and reduce the overall system cost. }\\
\hlcc{In future work, we aim  to address situations of partial beam measurements with our BeamsNet framework. Once there are less than three beams, the DVL does not provide an AUV velocity vector, and navigation solution  drifts. Specifically, we will examine the ability of this framework to regress the missing beams using DVL-only data and also examine the influence of using in addition to the inertial sensor readings. }

\section*{Acknowledgments}
N.C. is supported by the Maurice Hatter Foundation.

\bibliographystyle{cas-model2-names}

\newpage\bibliography{cas-refs}

\begin{thebibliography}{55}
\expandafter\ifx\csname natexlab\endcsname\relax\def\natexlab#1{#1}\fi
\providecommand{\url}[1]{\texttt{#1}}
\providecommand{\href}[2]{#2}
\providecommand{\path}[1]{#1}
\providecommand{\DOIprefix}{doi:}
\providecommand{\ArXivprefix}{arXiv:}
\providecommand{\URLprefix}{URL: }
\providecommand{\Pubmedprefix}{pmid:}
\providecommand{\doi}[1]{\href{http://dx.doi.org/#1}{\path{#1}}}
\providecommand{\Pubmed}[1]{\href{pmid:#1}{\path{#1}}}
\providecommand{\bibinfo}[2]{#2}
\ifx\xfnm\relax \def\xfnm[#1]{\unskip,\space#1}\fi
\bibitem[{Agarap(2018)}]{agarap2018deep}
\bibinfo{author}{Agarap, A.F.}, \bibinfo{year}{2018}.
\newblock \bibinfo{title}{Deep learning using rectified linear units (relu)}.
\newblock \bibinfo{journal}{arXiv preprint arXiv:1803.08375} .
\bibitem[{Ahmad et~al.(2013)Ahmad, Ghazilla, Khairi and
  Kasi}]{ahmad2013reviews}
\bibinfo{author}{Ahmad, N.}, \bibinfo{author}{Ghazilla, R.A.R.},
  \bibinfo{author}{Khairi, N.M.}, \bibinfo{author}{Kasi, V.},
  \bibinfo{year}{2013}.
\newblock \bibinfo{title}{Reviews on various inertial measurement unit (imu)
  sensor applications}.
\newblock \bibinfo{journal}{International Journal of Signal Processing Systems}
  \bibinfo{volume}{1}, \bibinfo{pages}{256--262}.
\bibitem[{Akeila et~al.(2013)Akeila, Salcic and Swain}]{akeila2013reducing}
\bibinfo{author}{Akeila, E.}, \bibinfo{author}{Salcic, Z.},
  \bibinfo{author}{Swain, A.}, \bibinfo{year}{2013}.
\newblock \bibinfo{title}{Reducing low-cost ins error accumulation in distance
  estimation using self-resetting}.
\newblock \bibinfo{journal}{IEEE Transactions on Instrumentation and
  Measurement} \bibinfo{volume}{63}, \bibinfo{pages}{177--184}.
\bibitem[{Armaghani and Asteris(2021)}]{armaghani2021comparative}
\bibinfo{author}{Armaghani, D.J.}, \bibinfo{author}{Asteris, P.G.},
  \bibinfo{year}{2021}.
\newblock \bibinfo{title}{A comparative study of ann and anfis models for the
  prediction of cement-based mortar materials compressive strength}.
\newblock \bibinfo{journal}{Neural Computing and Applications}
  \bibinfo{volume}{33}, \bibinfo{pages}{4501--4532}.
\bibitem[{Asraf et~al.(2021)Asraf, Shama and Klein}]{asraf2021pdrnet}
\bibinfo{author}{Asraf, O.}, \bibinfo{author}{Shama, F.},
  \bibinfo{author}{Klein, I.}, \bibinfo{year}{2021}.
\newblock \bibinfo{title}{Pdrnet: A deep-learning pedestrian dead reckoning
  framework}.
\newblock \bibinfo{journal}{IEEE Sensors Journal} .
\bibitem[{Bar-Shalom et~al.(2004)Bar-Shalom, Li and
  Kirubarajan}]{bar2004estimation}
\bibinfo{author}{Bar-Shalom, Y.}, \bibinfo{author}{Li, X.R.},
  \bibinfo{author}{Kirubarajan, T.}, \bibinfo{year}{2004}.
\newblock \bibinfo{title}{Estimation with applications to tracking and
  navigation: theory algorithms and software}.
\newblock \bibinfo{publisher}{John Wiley \& Sons}.
\bibitem[{Braginsky et~al.(2020)Braginsky, Baruch and
  Guterman}]{braginsky2020correction}
\bibinfo{author}{Braginsky, B.}, \bibinfo{author}{Baruch, A.},
  \bibinfo{author}{Guterman, H.}, \bibinfo{year}{2020}.
\newblock \bibinfo{title}{Correction of dvl error caused by seafloor gradient}.
\newblock \bibinfo{journal}{IEEE Sensors Journal} \bibinfo{volume}{20},
  \bibinfo{pages}{11652--11659}.
\bibitem[{Brokloff(1994)}]{brokloff1994matrix}
\bibinfo{author}{Brokloff, N.A.}, \bibinfo{year}{1994}.
\newblock \bibinfo{title}{Matrix algorithm for doppler sonar navigation}, in:
  \bibinfo{booktitle}{Proceedings of OCEANS'94}, \bibinfo{organization}{IEEE}.
  pp. \bibinfo{pages}{III--378}.
\bibitem[{Brossard et~al.(2020)Brossard, Bonnabel and
  Barrau}]{brossard2020denoising}
\bibinfo{author}{Brossard, M.}, \bibinfo{author}{Bonnabel, S.},
  \bibinfo{author}{Barrau, A.}, \bibinfo{year}{2020}.
\newblock \bibinfo{title}{Denoising imu gyroscopes with deep learning for
  open-loop attitude estimation}.
\newblock \bibinfo{journal}{IEEE Robotics and Automation Letters}
  \bibinfo{volume}{5}, \bibinfo{pages}{4796--4803}.
\bibitem[{Chen et~al.(2020)Chen, Zhao, Lu, Wang, Markham and
  Trigoni}]{chen2020deep}
\bibinfo{author}{Chen, C.}, \bibinfo{author}{Zhao, P.}, \bibinfo{author}{Lu,
  C.X.}, \bibinfo{author}{Wang, W.}, \bibinfo{author}{Markham, A.},
  \bibinfo{author}{Trigoni, N.}, \bibinfo{year}{2020}.
\newblock \bibinfo{title}{Deep-learning-based pedestrian inertial navigation:
  Methods, data set, and on-device inference}.
\newblock \bibinfo{journal}{IEEE Internet of Things Journal}
  \bibinfo{volume}{7}, \bibinfo{pages}{4431--4441}.
\bibitem[{Davari and Aguiar(2021)}]{davari2021real}
\bibinfo{author}{Davari, N.}, \bibinfo{author}{Aguiar, A.P.},
  \bibinfo{year}{2021}.
\newblock \bibinfo{title}{Real-time outlier detection applied to a doppler
  velocity log sensor based on hybrid autoencoder and recurrent neural
  network}.
\newblock \bibinfo{journal}{IEEE Journal of Oceanic Engineering}
  \bibinfo{volume}{46}, \bibinfo{pages}{1288--1301}.
\bibitem[{Eliav and Klein(2018)}]{eliav2018ins}
\bibinfo{author}{Eliav, R.}, \bibinfo{author}{Klein, I.}, \bibinfo{year}{2018}.
\newblock \bibinfo{title}{Ins/partial dvl measurements fusion with correlated
  process and measurement noise}, in: \bibinfo{booktitle}{Multidisciplinary
  Digital Publishing Institute Proceedings}, p.~\bibinfo{pages}{34}.
\bibitem[{Gonzalez(2018)}]{gonzalez2018deep}
\bibinfo{author}{Gonzalez, R.C.}, \bibinfo{year}{2018}.
\newblock \bibinfo{title}{Deep convolutional neural networks [lecture notes]}.
\newblock \bibinfo{journal}{IEEE Signal Processing Magazine}
  \bibinfo{volume}{35}, \bibinfo{pages}{79--87}.
\bibitem[{Groves(2015)}]{groves2015principles}
\bibinfo{author}{Groves, P.D.}, \bibinfo{year}{2015}.
\newblock \bibinfo{title}{Principles of gnss, inertial, and multisensor
  integrated navigation systems}.
\newblock \bibinfo{journal}{IEEE Aerospace and Electronic Systems Magazine}
  \bibinfo{volume}{30}, \bibinfo{pages}{26--27}.
\bibitem[{Gu et~al.(2018)Gu, Khoshelham, Yu and Shang}]{gu2018accurate}
\bibinfo{author}{Gu, F.}, \bibinfo{author}{Khoshelham, K.},
  \bibinfo{author}{Yu, C.}, \bibinfo{author}{Shang, J.}, \bibinfo{year}{2018}.
\newblock \bibinfo{title}{Accurate step length estimation for pedestrian dead
  reckoning localization using stacked autoencoders}.
\newblock \bibinfo{journal}{IEEE Transactions on Instrumentation and
  Measurement} \bibinfo{volume}{68}, \bibinfo{pages}{2705--2713}.
\bibitem[{Han et~al.(2021)Han, Meng, Zhang and Yan}]{han2021hybrid}
\bibinfo{author}{Han, S.}, \bibinfo{author}{Meng, Z.}, \bibinfo{author}{Zhang,
  X.}, \bibinfo{author}{Yan, Y.}, \bibinfo{year}{2021}.
\newblock \bibinfo{title}{Hybrid deep recurrent neural networks for noise
  reduction of mems-imu with static and dynamic conditions}.
\newblock \bibinfo{journal}{Micromachines} \bibinfo{volume}{12},
  \bibinfo{pages}{214}.
\bibitem[{He et~al.(2015)He, Zhang, Ren and Sun}]{he2015delving}
\bibinfo{author}{He, K.}, \bibinfo{author}{Zhang, X.}, \bibinfo{author}{Ren,
  S.}, \bibinfo{author}{Sun, J.}, \bibinfo{year}{2015}.
\newblock \bibinfo{title}{Delving deep into rectifiers: Surpassing human-level
  performance on imagenet classification}, in: \bibinfo{booktitle}{Proceedings
  of the IEEE international conference on computer vision}, pp.
  \bibinfo{pages}{1026--1034}.
\bibitem[{Hu et~al.(2021)Hu, Zhang, Tan, Ruan, Agia and Nejat}]{hu2021sim}
\bibinfo{author}{Hu, H.}, \bibinfo{author}{Zhang, K.}, \bibinfo{author}{Tan,
  A.H.}, \bibinfo{author}{Ruan, M.}, \bibinfo{author}{Agia, C.},
  \bibinfo{author}{Nejat, G.}, \bibinfo{year}{2021}.
\newblock \bibinfo{title}{A sim-to-real pipeline for deep reinforcement
  learning for autonomous robot navigation in cluttered rough terrain}.
\newblock \bibinfo{journal}{IEEE Robotics and Automation Letters}
  \bibinfo{volume}{6}, \bibinfo{pages}{6569--6576}.
\bibitem[{Jain et~al.(2015)Jain, Mohammad, Bora and Singh}]{jain2015review}
\bibinfo{author}{Jain, S.K.}, \bibinfo{author}{Mohammad, S.},
  \bibinfo{author}{Bora, S.}, \bibinfo{author}{Singh, M.},
  \bibinfo{year}{2015}.
\newblock \bibinfo{title}{A review paper on: autonomous underwater vehicle}.
\newblock \bibinfo{journal}{International Journal of Scientific \& Engineering
  Research} \bibinfo{volume}{6}, \bibinfo{pages}{38}.
\bibitem[{Klein(2021)}]{klein2021ins}
\bibinfo{author}{Klein, I.}, \bibinfo{year}{2021}.
\newblock \bibinfo{title}{Ins drift mitigation during dvl outages}, in:
  \bibinfo{booktitle}{OCEANS 2021: San Diego--Porto},
  \bibinfo{organization}{IEEE}. pp. \bibinfo{pages}{1--5}.
\bibitem[{Leonard and Bahr(2016)}]{leonard2016autonomous}
\bibinfo{author}{Leonard, J.J.}, \bibinfo{author}{Bahr, A.},
  \bibinfo{year}{2016}.
\newblock \bibinfo{title}{Autonomous underwater vehicle navigation}.
\newblock \bibinfo{journal}{Springer handbook of ocean engineering} ,
  \bibinfo{pages}{341--358}.
\bibitem[{Li et~al.(2021)Li, Xu, He and Wu}]{li2021underwater}
\bibinfo{author}{Li, D.}, \bibinfo{author}{Xu, J.}, \bibinfo{author}{He, H.},
  \bibinfo{author}{Wu, M.}, \bibinfo{year}{2021}.
\newblock \bibinfo{title}{An underwater integrated navigation algorithm to deal
  with dvl malfunctions based on deep learning}.
\newblock \bibinfo{journal}{IEEE Access} \bibinfo{volume}{9},
  \bibinfo{pages}{82010--82020}.
\bibitem[{Li et~al.(2015)Li, Zhang, Sun, Yang, Chen and Li}]{li2015alignment}
\bibinfo{author}{Li, W.}, \bibinfo{author}{Zhang, L.}, \bibinfo{author}{Sun,
  F.}, \bibinfo{author}{Yang, L.}, \bibinfo{author}{Chen, M.},
  \bibinfo{author}{Li, Y.}, \bibinfo{year}{2015}.
\newblock \bibinfo{title}{Alignment calibration of imu and doppler sensors for
  precision ins/dvl integrated navigation}.
\newblock \bibinfo{journal}{Optik} \bibinfo{volume}{126},
  \bibinfo{pages}{3872--3876}.
\bibitem[{Liu et~al.(2018a)Liu, Wang, Deng and Fu}]{liu2018ins}
\bibinfo{author}{Liu, P.}, \bibinfo{author}{Wang, B.}, \bibinfo{author}{Deng,
  Z.}, \bibinfo{author}{Fu, M.}, \bibinfo{year}{2018}a.
\newblock \bibinfo{title}{Ins/dvl/ps tightly coupled underwater navigation
  method with limited dvl measurements}.
\newblock \bibinfo{journal}{IEEE Sensors Journal} \bibinfo{volume}{18},
  \bibinfo{pages}{2994--3002}.
\bibitem[{Liu et~al.(2022)Liu, Wang, Li, Hou, Zhu and Wang}]{liu2022sins}
\bibinfo{author}{Liu, P.}, \bibinfo{author}{Wang, B.}, \bibinfo{author}{Li,
  G.}, \bibinfo{author}{Hou, D.}, \bibinfo{author}{Zhu, Z.},
  \bibinfo{author}{Wang, Z.}, \bibinfo{year}{2022}.
\newblock \bibinfo{title}{Sins/dvl integrated navigation method with current
  compensation using rbf neural network}.
\newblock \bibinfo{journal}{IEEE Sensors Journal} .
\bibitem[{Liu et~al.(2021)Liu, Liu, Liu and Zhang}]{liu2021modified}
\bibinfo{author}{Liu, R.}, \bibinfo{author}{Liu, F.}, \bibinfo{author}{Liu,
  C.}, \bibinfo{author}{Zhang, P.}, \bibinfo{year}{2021}.
\newblock \bibinfo{title}{Modified sage-husa adaptive kalman filter-based
  sins/dvl integrated navigation system for auv}.
\newblock \bibinfo{journal}{Journal of Sensors} \bibinfo{volume}{2021}.
\bibitem[{Liu et~al.(2018b)Liu, Fan, Lv, Wu, Li and Ding}]{liu2018innovative}
\bibinfo{author}{Liu, Y.}, \bibinfo{author}{Fan, X.}, \bibinfo{author}{Lv, C.},
  \bibinfo{author}{Wu, J.}, \bibinfo{author}{Li, L.}, \bibinfo{author}{Ding,
  D.}, \bibinfo{year}{2018}b.
\newblock \bibinfo{title}{An innovative information fusion method with adaptive
  kalman filter for integrated ins/gps navigation of autonomous vehicles}.
\newblock \bibinfo{journal}{Mechanical Systems and Signal Processing}
  \bibinfo{volume}{100}, \bibinfo{pages}{605--616}.
\bibitem[{Lv et~al.(2021)Lv, He and Guo}]{lv2021position}
\bibinfo{author}{Lv, P.F.}, \bibinfo{author}{He, B.}, \bibinfo{author}{Guo,
  J.}, \bibinfo{year}{2021}.
\newblock \bibinfo{title}{Position correction model based on gated hybrid rnn
  for auv navigation}.
\newblock \bibinfo{journal}{IEEE Transactions on Vehicular Technology}
  \bibinfo{volume}{70}, \bibinfo{pages}{5648--5657}.
\bibitem[{Lv et~al.(2020)Lv, He, Guo, Shen, Yan and Sha}]{lv2020underwater}
\bibinfo{author}{Lv, P.F.}, \bibinfo{author}{He, B.}, \bibinfo{author}{Guo,
  J.}, \bibinfo{author}{Shen, Y.}, \bibinfo{author}{Yan, T.H.},
  \bibinfo{author}{Sha, Q.X.}, \bibinfo{year}{2020}.
\newblock \bibinfo{title}{Underwater navigation methodology based on
  intelligent velocity model for standard auv}.
\newblock \bibinfo{journal}{Ocean Engineering} \bibinfo{volume}{202},
  \bibinfo{pages}{107073}.
\bibitem[{Manalang et~al.(2018)Manalang, Delaney, Marburg and
  Nawaz}]{manalang2018resident}
\bibinfo{author}{Manalang, D.}, \bibinfo{author}{Delaney, J.},
  \bibinfo{author}{Marburg, A.}, \bibinfo{author}{Nawaz, A.},
  \bibinfo{year}{2018}.
\newblock \bibinfo{title}{Resident auv workshop 2018: Applications and a path
  forward}, in: \bibinfo{booktitle}{2018 IEEE/OES Autonomous Underwater Vehicle
  Workshop (AUV)}, \bibinfo{organization}{IEEE}. pp. \bibinfo{pages}{1--6}.
\bibitem[{Mu et~al.(2019)Mu, He, Zhang, Song, Shen and Feng}]{mu2019end}
\bibinfo{author}{Mu, X.}, \bibinfo{author}{He, B.}, \bibinfo{author}{Zhang,
  X.}, \bibinfo{author}{Song, Y.}, \bibinfo{author}{Shen, Y.},
  \bibinfo{author}{Feng, C.}, \bibinfo{year}{2019}.
\newblock \bibinfo{title}{End-to-end navigation for autonomous underwater
  vehicle with hybrid recurrent neural networks}.
\newblock \bibinfo{journal}{Ocean Engineering} \bibinfo{volume}{194},
  \bibinfo{pages}{106602}.
\bibitem[{Myung(2003)}]{myung2003tutorial}
\bibinfo{author}{Myung, I.J.}, \bibinfo{year}{2003}.
\newblock \bibinfo{title}{Tutorial on maximum likelihood estimation}.
\newblock \bibinfo{journal}{Journal of mathematical Psychology}
  \bibinfo{volume}{47}, \bibinfo{pages}{90--100}.
\bibitem[{Nicholson and Healey(2008)}]{nicholson2008present}
\bibinfo{author}{Nicholson, J.}, \bibinfo{author}{Healey, A.},
  \bibinfo{year}{2008}.
\newblock \bibinfo{title}{The present state of autonomous underwater vehicle
  (auv) applications and technologies}.
\newblock \bibinfo{journal}{Marine Technology Society Journal}
  \bibinfo{volume}{42}, \bibinfo{pages}{44--51}.
\bibitem[{Paull et~al.(2013)Paull, Saeedi, Seto and Li}]{paull2013auv}
\bibinfo{author}{Paull, L.}, \bibinfo{author}{Saeedi, S.},
  \bibinfo{author}{Seto, M.}, \bibinfo{author}{Li, H.}, \bibinfo{year}{2013}.
\newblock \bibinfo{title}{Auv navigation and localization: A review}.
\newblock \bibinfo{journal}{IEEE Journal of oceanic engineering}
  \bibinfo{volume}{39}, \bibinfo{pages}{131--149}.
\bibitem[{Ruder(2016)}]{ruder2016overview}
\bibinfo{author}{Ruder, S.}, \bibinfo{year}{2016}.
\newblock \bibinfo{title}{An overview of gradient descent optimization
  algorithms}.
\newblock \bibinfo{journal}{arXiv preprint arXiv:1609.04747} .
\bibitem[{Saksvik et~al.(2021)Saksvik, Alcocer and Hassani}]{saksvik2021deep}
\bibinfo{author}{Saksvik, I.B.}, \bibinfo{author}{Alcocer, A.},
  \bibinfo{author}{Hassani, V.}, \bibinfo{year}{2021}.
\newblock \bibinfo{title}{A deep learning approach to dead-reckoning navigation
  for autonomous underwater vehicles with limited sensor payloads}, in:
  \bibinfo{booktitle}{OCEANS 2021: San Diego--Porto},
  \bibinfo{organization}{IEEE}. pp. \bibinfo{pages}{1--9}.
\bibitem[{Shalev and Klein(2021)}]{shalev2021botnet}
\bibinfo{author}{Shalev, H.}, \bibinfo{author}{Klein, I.},
  \bibinfo{year}{2021}.
\newblock \bibinfo{title}{Botnet: Deep learning-based bearings-only tracking
  using multiple passive sensors}.
\newblock \bibinfo{journal}{Sensors} \bibinfo{volume}{21},
  \bibinfo{pages}{4457}.
\bibitem[{Sharma et~al.(2017)Sharma, Sharma and Athaiya}]{sharma2017activation}
\bibinfo{author}{Sharma, S.}, \bibinfo{author}{Sharma, S.},
  \bibinfo{author}{Athaiya, A.}, \bibinfo{year}{2017}.
\newblock \bibinfo{title}{Activation functions in neural networks}.
\newblock \bibinfo{journal}{towards data science} \bibinfo{volume}{6},
  \bibinfo{pages}{310--316}.
\bibitem[{Shin and El-Sheimy(2002)}]{shin2002accuracy}
\bibinfo{author}{Shin, E.H.}, \bibinfo{author}{El-Sheimy, N.},
  \bibinfo{year}{2002}.
\newblock \bibinfo{title}{Accuracy improvement of low cost ins/gps for land
  applications}, in: \bibinfo{booktitle}{Proceedings of the 2002 national
  technical meeting of the institute of navigation}, pp.
  \bibinfo{pages}{146--157}.
\bibitem[{Shurin et~al.(2022)Shurin, Saraev, Yona, Gutnik, Faber, Etzion and
  Klein}]{shurin2022autonomous}
\bibinfo{author}{Shurin, A.}, \bibinfo{author}{Saraev, A.},
  \bibinfo{author}{Yona, M.}, \bibinfo{author}{Gutnik, Y.},
  \bibinfo{author}{Faber, S.}, \bibinfo{author}{Etzion, A.},
  \bibinfo{author}{Klein, I.}, \bibinfo{year}{2022}.
\newblock \bibinfo{title}{The autonomous platforms inertial dataset}.
\newblock \bibinfo{journal}{IEEE Access} \bibinfo{volume}{10},
  \bibinfo{pages}{10191--10201}.
\bibitem[{Simon(2006)}]{simon2006optimal}
\bibinfo{author}{Simon, D.}, \bibinfo{year}{2006}.
\newblock \bibinfo{title}{Optimal state estimation: Kalman, H infinity, and
  nonlinear approaches}.
\newblock \bibinfo{publisher}{John Wiley \& Sons}.
\bibitem[{Sohn and Kim(1997)}]{sohn1997detection}
\bibinfo{author}{Sohn, B.Y.}, \bibinfo{author}{Kim, G.B.},
  \bibinfo{year}{1997}.
\newblock \bibinfo{title}{Detection of outliers in weighted least squares
  regression}.
\newblock \bibinfo{journal}{Korean Journal of Computational \& Applied
  Mathematics} \bibinfo{volume}{4}, \bibinfo{pages}{441--452}.
\bibitem[{Stoica and Nehorai(1989)}]{stoica1989music}
\bibinfo{author}{Stoica, P.}, \bibinfo{author}{Nehorai, A.},
  \bibinfo{year}{1989}.
\newblock \bibinfo{title}{Music, maximum likelihood, and cramer-rao bound}.
\newblock \bibinfo{journal}{IEEE Transactions on Acoustics, speech, and signal
  processing} \bibinfo{volume}{37}, \bibinfo{pages}{720--741}.
\bibitem[{Tal et~al.(2017)Tal, Klein and Katz}]{tal2017inertial}
\bibinfo{author}{Tal, A.}, \bibinfo{author}{Klein, I.}, \bibinfo{author}{Katz,
  R.}, \bibinfo{year}{2017}.
\newblock \bibinfo{title}{Inertial navigation system/doppler velocity log
  (ins/dvl) fusion with partial dvl measurements}.
\newblock \bibinfo{journal}{Sensors} \bibinfo{volume}{17},
  \bibinfo{pages}{415}.
\bibitem[{Teledyne()}]{Teledyne}
\bibinfo{author}{Teledyne}, .
\newblock \bibinfo{title}{{Teledyne Marine RD Instruments DVL}}.
\newblock \bibinfo{howpublished}{Available:
  \url{http://www.teledynemarine.com/dvls}}.
\newblock \bibinfo{note}{Accessed: 2021-07-1}.
\bibitem[{Thong et~al.(2002)Thong, Woolfson, Crowe, Hayes-Gill and
  Challis}]{thong2002dependence}
\bibinfo{author}{Thong, Y.}, \bibinfo{author}{Woolfson, M.},
  \bibinfo{author}{Crowe, J.}, \bibinfo{author}{Hayes-Gill, B.},
  \bibinfo{author}{Challis, R.}, \bibinfo{year}{2002}.
\newblock \bibinfo{title}{Dependence of inertial measurements of distance on
  accelerometer noise}.
\newblock \bibinfo{journal}{Measurement Science and Technology}
  \bibinfo{volume}{13}, \bibinfo{pages}{1163}.
\bibitem[{Titterton et~al.(2004)Titterton, Weston and
  Weston}]{titterton2004strapdown}
\bibinfo{author}{Titterton, D.}, \bibinfo{author}{Weston, J.L.},
  \bibinfo{author}{Weston, J.}, \bibinfo{year}{2004}.
\newblock \bibinfo{title}{Strapdown inertial navigation technology}.
  volume~\bibinfo{volume}{17}.
\newblock \bibinfo{publisher}{IET}.
\bibitem[{Topini et~al.(2020)Topini, Topini, Franchi, Bucci, Secciani, Ridolfi
  and Allotta}]{topini2020lstm}
\bibinfo{author}{Topini, E.}, \bibinfo{author}{Topini, A.},
  \bibinfo{author}{Franchi, M.}, \bibinfo{author}{Bucci, A.},
  \bibinfo{author}{Secciani, N.}, \bibinfo{author}{Ridolfi, A.},
  \bibinfo{author}{Allotta, B.}, \bibinfo{year}{2020}.
\newblock \bibinfo{title}{Lstm-based dead reckoning navigation for autonomous
  underwater vehicles}, in: \bibinfo{booktitle}{Global Oceans 2020:
  Singapore--US Gulf Coast}, \bibinfo{organization}{IEEE}. pp.
  \bibinfo{pages}{1--7}.
\bibitem[{Wang et~al.(2019)Wang, Xu, Yao, Zhang and Zhu}]{wang2019novel}
\bibinfo{author}{Wang, D.}, \bibinfo{author}{Xu, X.}, \bibinfo{author}{Yao,
  Y.}, \bibinfo{author}{Zhang, T.}, \bibinfo{author}{Zhu, Y.},
  \bibinfo{year}{2019}.
\newblock \bibinfo{title}{A novel sins/dvl tightly integrated navigation method
  for complex environment}.
\newblock \bibinfo{journal}{IEEE Transactions on Instrumentation and
  Measurement} \bibinfo{volume}{69}, \bibinfo{pages}{5183--5196}.
\bibitem[{Yona and Klein(2021)}]{yona2021compensating}
\bibinfo{author}{Yona, M.}, \bibinfo{author}{Klein, I.}, \bibinfo{year}{2021}.
\newblock \bibinfo{title}{Compensating for partial doppler velocity log outages
  by using deep-learning approaches}, in: \bibinfo{booktitle}{2021 IEEE
  International Symposium on Robotic and Sensors Environments (ROSE)},
  \bibinfo{organization}{IEEE}. pp. \bibinfo{pages}{1--5}.
\bibitem[{Zhang et~al.(2018)Zhang, Fei, Zhu, Mu, Lv, Liu, He and
  Yan}]{zhang2018novel}
\bibinfo{author}{Zhang, X.}, \bibinfo{author}{Fei, X.}, \bibinfo{author}{Zhu,
  Y.}, \bibinfo{author}{Mu, X.}, \bibinfo{author}{Lv, P.},
  \bibinfo{author}{Liu, H.}, \bibinfo{author}{He, B.}, \bibinfo{author}{Yan,
  T.}, \bibinfo{year}{2018}.
\newblock \bibinfo{title}{Novel improved ukf algorithm and its application in
  auv navigation system}, in: \bibinfo{booktitle}{2018 OCEANS-MTS/IEEE Kobe
  Techno-Oceans (OTO)}, \bibinfo{organization}{IEEE}. pp.
  \bibinfo{pages}{1--4}.
\bibitem[{Zhang et~al.(2020)Zhang, He, Li, Mu, Zhou and Mang}]{zhang2020navnet}
\bibinfo{author}{Zhang, X.}, \bibinfo{author}{He, B.}, \bibinfo{author}{Li,
  G.}, \bibinfo{author}{Mu, X.}, \bibinfo{author}{Zhou, Y.},
  \bibinfo{author}{Mang, T.}, \bibinfo{year}{2020}.
\newblock \bibinfo{title}{Navnet: Auv navigation through deep sequential
  learning}.
\newblock \bibinfo{journal}{IEEE Access} \bibinfo{volume}{8},
  \bibinfo{pages}{59845--59861}.
\bibitem[{Zhang et~al.(2019)Zhang, Mu, Liu, He and Yan}]{zhang2019application}
\bibinfo{author}{Zhang, X.}, \bibinfo{author}{Mu, X.}, \bibinfo{author}{Liu,
  H.}, \bibinfo{author}{He, B.}, \bibinfo{author}{Yan, T.},
  \bibinfo{year}{2019}.
\newblock \bibinfo{title}{Application of modified ekf based on intelligent data
  fusion in auv navigation}, in: \bibinfo{booktitle}{2019 IEEE Underwater
  Technology (UT)}, \bibinfo{organization}{IEEE}. pp. \bibinfo{pages}{1--4}.
\bibitem[{Zhao et~al.(2017)Zhao, Lu, Chen, Liu and Wu}]{zhao2017convolutional}
\bibinfo{author}{Zhao, B.}, \bibinfo{author}{Lu, H.}, \bibinfo{author}{Chen,
  S.}, \bibinfo{author}{Liu, J.}, \bibinfo{author}{Wu, D.},
  \bibinfo{year}{2017}.
\newblock \bibinfo{title}{Convolutional neural networks for time series
  classification}.
\newblock \bibinfo{journal}{Journal of Systems Engineering and Electronics}
  \bibinfo{volume}{28}, \bibinfo{pages}{162--169}.
\bibitem[{Zhu et~al.(2017)Zhu, Mottaghi, Kolve, Lim, Gupta, Fei-Fei and
  Farhadi}]{zhu2017target}
\bibinfo{author}{Zhu, Y.}, \bibinfo{author}{Mottaghi, R.},
  \bibinfo{author}{Kolve, E.}, \bibinfo{author}{Lim, J.J.},
  \bibinfo{author}{Gupta, A.}, \bibinfo{author}{Fei-Fei, L.},
  \bibinfo{author}{Farhadi, A.}, \bibinfo{year}{2017}.
\newblock \bibinfo{title}{Target-driven visual navigation in indoor scenes
  using deep reinforcement learning}, in: \bibinfo{booktitle}{2017 IEEE
  international conference on robotics and automation (ICRA)},
  \bibinfo{organization}{IEEE}. pp. \bibinfo{pages}{3357--3364}.

\end{thebibliography}

\end{document}